\pdfoutput=1

\documentclass[11pt]{article}

\usepackage{acl}

\usepackage{times}
\usepackage{latexsym}
\usepackage{microtype}
\usepackage{graphicx}
\usepackage{verbatim}
\usepackage{booktabs} 

\usepackage{hyperref}

\usepackage{multirow}
\usepackage{tabularx}
\usepackage{colortbl}
\usepackage{booktabs}
\usepackage{adjustbox}

\usepackage{amsmath}
\usepackage{amssymb}
\usepackage{caption}
\usepackage{float}

\usepackage{mathtools}
\usepackage{amsthm}
\usepackage{listings}
\usepackage{xcolor}

\usepackage{todonotes}
\usepackage{subcaption}
\usepackage{enumitem}
\usepackage[capitalize,noabbrev]{cleveref}

\usepackage[T1]{fontenc}

\usepackage[utf8]{inputenc}

\usepackage{microtype}

\usepackage{inconsolata}

\title{Fortran2CPP: Automating Fortran-to-C++ Translation using LLMs via Multi-Turn Dialogue and Dual-Agent Integration}

\author{Le Chen*$^{1}$ \quad Bin Lei*$^{2}$ \quad Dunzhi Zhou$^{3}$ \\ {\bf Pei-Hung Lin}$^{4}$ 
\quad {\bf Chunhua Liao}$^{4}$ \quad {\bf Caiwen Ding}$^{2}$ \quad {\bf Ali Jannesari}$^{1}$
\\\\ 
\textsuperscript{1}Iowa State University, \textsuperscript{2}University of Minnesota, \textsuperscript{3}North Carolina State University,\\
\textsuperscript{4}Lawrence Livermore National Laboratory
\thanks{* These authors contributed equally to this work. This work was performed under the auspices of the U.S. Department of Energy by Lawrence Livermore National Laboratory under Contract DE-AC52-07NA27344, also supported by the U.S. Dept. of Energy, Office of Science, Advanced Scientific Computing Program (ASCR SC-21). LLNL-CONF-870817.}
}

\begin{document}
\maketitle
\begin{abstract}

Translating legacy Fortran code into C++ is a crucial step in modernizing high-performance computing (HPC) applications. However, the scarcity of high-quality, parallel Fortran-to-C++ datasets and the limited domain-specific expertise in large language models (LLMs) present significant challenges for automated translation. In this paper, we introduce Fortran2CPP, a multi-turn dialogue dataset generated by a novel LLM agent-based approach that integrates a dual-LLM Questioner-Solver module to enhance translation accuracy. 
Our dataset comprises 11.7k dialogues capturing iterative feedback-decision workflows including code translation,  compilation, execution, unit testing, and error-fixing. Using this dataset, we fine-tune several open-weight LLMs and achieve up to a 3.31× improvement in CodeBLEU scores and a 92\% increase in compilation success rate, demonstrating enhanced syntactic accuracy and functional reliability. Our findings highlight the value of dialogue-based LLM training for complex code translation tasks. The dataset and model have been open-sourced and are available on our public GitHub repository\footnote{\url{https://github.com/HPC-Fortran2CPP/Fortran2Cpp}}. 

\end{abstract}

\section{Introduction}

Translating legacy Fortran code into C++ has become a crucial strategy in high-performance computing (HPC) to modernize projects, enhance maintainability, and improve performance~\cite{czarnul2020survey}. 
Traditional algorithm-based code translation approaches, which rely on meticulously crafted rules and patterns and a deep understanding of source and target languages semantics and logic, often incur high development and maintenance costs with limited flexibility. To address these challenges, researchers have turned to machine learning-based approaches \cite{roziere2020unsupervised, roziere2021leveraging, szafraniec2022code} for more adaptable and effective translations. Recent advancements in Large Language Models (LLMs)
and their successful in code related tasks such as code completion \citep{zhang2024llm}, parallelization \cite{chen2024ompgpt}, and code documentation generation  \citep{luo2024repoagent}, have further spurred interest in leveraging LLMs for code translation.

However, empirical studies reveal that neither general-purpose nor code-specific LLMs can reliably automate code translation across various programming languages, including C, C++, Go, Java, and Python~\citep{pan2024lost}. For instance, models like CodeLlama-13B, StarCoder, and GPT-6.7B achieve average manual assessment scores of 4.53/5, Pass@1 scores of 0.61 in execution testing, and CodeBLEU scores of 0.192. Even GPT-4, despite its advanced capabilities, shows only marginal improvements, with scores of 4.575/5, 0.655, and 0.232, respectively. These results underscore the limitations of current LLMs in code translation tasks, particularly for Fortran-to-C++ translation.

The suboptimal performance of LLMs in this domain can be attributed to two key factors. First, the lack of open-source Fortran data severely limits the models' ability to develop a robust \textbf{understanding of low-resource languages like Fortran}. An analysis of GitHub data shows that only 0.04\% of code is written in Fortran\footnote{\url{https://madnight.github.io/githut/\#/pull_requests/2024/1}}, underscoring the data scarcity issue. Second, \textbf{the lack of high-quality paired Fortran-C++ datasets} makes it challenging to fine-tune or supervise-train models to learn the nuanced patterns required for accurate translation. Previous efforts, such as \citet{lei2023creating}, attempted to address this by merging existing HPC datasets. However, the resulting dataset still lacked sufficient Fortran knowledge and was limited in size, highlighting the need for more comprehensive and specialized resources.

To overcome these challenges, we introduce Fortran2CPP with an LLM agent-based approach specifically designed for Fortran-to-C++ translation. Starting from public Fortran code, our approach incorporates custom scripts and tools to enable more accurate and efficient translations of complex Fortran code. The novel Questioner-Solver module enables construction of a dialogue dataset that captures iterative feedback-decision cycles, validation results, and detailed error messages, providing rich insight for LLMs. This dataset not only contains verified Fortran CPP code pairs but also serves as a valuable domain specific knowledge resource for training and fine-tuning future models.

Our contributions are as follows:

\begin{itemize}

\item \textbf{An innovative LLM agent-based approach for Fortran-C++ translation:} We introduce an LLM agent-based approach that automatically incorporates various verification processes in iterative loops for Fortran-C++ translation, requiring minimal human intervention. This approach is scalable for dataset generation by taking incrementally larger seed Fortran code sets from GitHub to generate equivalent translated C++ programs.

\item \textbf{The Questioner-Solver module design:} Our novel Questioner-Solver module design advances beyond agents with a single LLM by offloading referencing and decision-making tasks to separate LLMs. Operating in iterative loops, this module tracks inference, output, verification results, and solutions. The resulting process dialogue effectively extends the knowledge of LLMs in low-resource languages such as Fortran.

\item \textbf{A Multi-turn dialogue dataset to support LLMs in Fortran-C++ translation:} We generate a multi-turn dialogue dataset that captures the iterative feedback-decision cycles, validation results, and detailed error messages during the translation. This dataset not only includes verified Fortran-C++ code pairs but also provides rich insights into the reasoning steps and error-fixing workflows. This comprehensive approach not only improves translation accuracy but also provides rich insights into the reasoning process, serving as an invaluable resource for training and fine-tuning future models in Fortran-C++ translation tasks. Ultimately, generating dialogue datasets is more efficient than the process of creating traditional code pair datasets.


\item \textbf{Comprehensive evaluation:} By fine-tuning on our dialogue dataset, the one-shot code translation capabilities of three models, DeepSeek-Coder (6.7B), CodeLlama (13B), and StarCoder (15.5B), have been significantly enhanced, achieving achieving a 1.5x to 3.3x increase in their CodeBLEU scores. This demonstrates the effectiveness of the dialogue dataset in improving LLMs' performance in low-resource languages. Moreover, we extended the HumanEval dataset by contributing the Fortran version data for evaluation.
\end{itemize}

\section{Background}
This section explores the background of Fortran to C++ translation and discusses the current advancements and associated challenges for this purpose.



\subsection{Fortran to C++ Translation}
Translating Fortran to C++ is crucial for modernizing legacy scientific programs. Early efforts relied on manual expert-driven interfaces~\citep{gray1999shadow, morris2012exploring}. Recent studies have shifted towards automated techniques using glue code and intermediate representations~\citep{seragiotto2004standardized, johnson2019automated, grosse2012automatic}. However, these methods often have limited applicability and still require expensive manual adaptation to evolving programming languages.


\subsection{Challenges in Employing LLMs for Fortran to C++ Translation}
\label{sec:bkg-chanllenge}
LLMs have shown promise in HPC~\citep{chen2023lm4hpc,ding2023hpc,chen2023data} and programming language translation~\citep{yang2024exploring}. However, applying LLMs to Fortran-C++ translation faces challenges due to limited datasets for fine-tuning and evaluation. Beyond standard code snippet pairs, there's a need for diverse datasets, including multi-turn dialogues capturing the translation process with compilation and runtime feedback. Developing tailored evaluation methods is also crucial for accurate model assessment.

\subsection{LLM Agent System}
The swift progress of large language models has generated considerable enthusiasm for utilizing them to tackle intricate, real-world challenges. However, even with their achievements, LLMs frequently struggle with tasks that necessitate multiple steps or more profound analysis~\citep{guo2024large}. LLM agents present a strong solution to this issue by integrating effective reasoning, memory, and tool utilization. An LLM agent system can be described as a computational framework that harnesses the reasoning, planning, and execution abilities of a large language model.  

\section{Approach}
This section introduces a novel Questioner-Solver module and a multi-phase LLM agent pipeline to create a high-quality Fortran-C++ dataset. The approach is designed to handle the complexities of Fortran code, ensure functional equivalence in the translated C++ code, and generate a multi-turn dialogue dataset for training and fine-tuning LLMs.

\subsection{The Questioner-Solver Module}
\label{sec:approach-module}

\begin{figure}[h]
\centering
\includegraphics[width=0.48\textwidth]{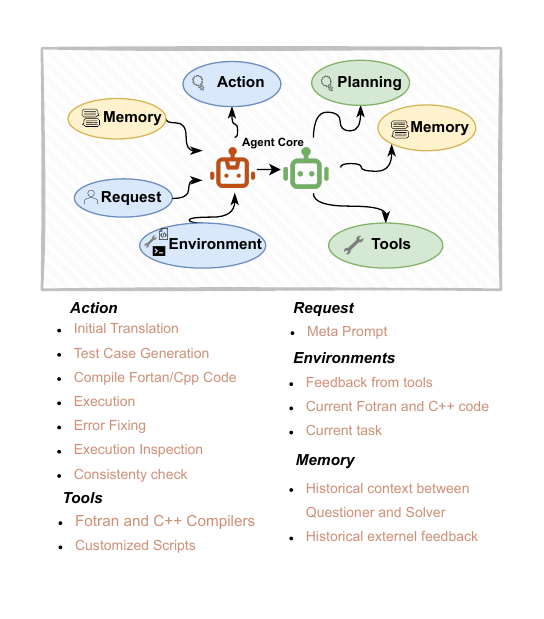} 
\caption{The questioner-solver model serves as the core of LLM agents in our approach.}
\label{fig:agent}
\end{figure}


With the development of LLM agents, generating data using LLM agent-based systems has broad application potential~\cite{wu2023autogen, toubal2024modeling}. These system usually leverage various LLM agents to verify and refine the data in iterative loops. In Section~\ref{sec:pipeline_detail}, we introduce our pipeline for generating Fortran-CPP data by translating seed Fortran code into equivalent C++ code. Unlike previous LLM agent-based code translation approaches, our Fortran2CPP dataset captures multi-turn dialogues that include intermediate information such as incorrect translations, error messages, and error-fixing steps. This is achieved through the core of our approach: the Questioner-Solver module, which employs two LLMs instead of one as the central component of the LLM agent system.

As shown in Figure~\ref{fig:agent}, the Questioner assesses the current state and formulates pertinent questions for the Solver, which then responds and determines subsequent actions. By dividing the agent's responsibilities between two components, the Questioner dynamically generates questions that incorporate essential information from the current memory state and environmental tool feedback. The Solver, on the other hand, is responsible for planning and execution tasks, including translation, error correction, and the invocation of tools or scripts.




The Questioner-Solver design offers several significant advantages over a single-LLM approach:
\begin{itemize}[noitemsep, topsep=0pt, leftmargin=*]
    \item Expert-like Reasoning: The design mimics expert problem-solving by dividing the task into two specialized roles: environment assessment (Questioner) and decision-making (Solver). This separation allows for a more nuanced approach to complex problems, particularly in code translation tasks.

    \item Increased Autonomy: The design significantly reduces the need for user intervention or manual Python coding to process different LLM outputs. This autonomy enables the system to operate continuously and independently, making it more scalable for large-scale dataset generation.

    \item Domain-Specific Expertise: The Questioner-Solver interaction facilitates a rich, knowledge-driven dialogue along the translation pipeline. The tracked interaction history accumulates valuable domain-specific knowledge, encompassing processes such as Fortran-to-C++ translation, multi-stage correctness verification, and error correction. This accumulated expertise not only enhances the system's performance but also serves as a valuable dataset for fine-tuning LLMs on low-resource programming languages like Fortran.

    \item Adaptive Problem-Solving: The iterative nature of the Questioner-Solver interaction allows for dynamic adaptation to evolving challenges, making it particularly useful in complex coding scenarios where errors and corrections are frequent.   
\end{itemize}

Section~\ref{sec:pipeline_detail} discusses our pipeline for translating Fortran to C++ and generating multi-turn dialogue data. At each step, the Questioner-Solver module handles the dynamic, uncertain, and complex environment. The process operates over a sequence of time steps $t=1,...,T$. At each time step $t$:

Questioner: The Questioner analyzes the current memory state, $mem_t$, and the environmental context, $env_{t}$, to evaluate the system's current status. Based on this assessment, the Questioner determines an appropriate action, $act_t$, and formulates a corresponding question, $qes_t$, guided by a set of example prompts, $plist_{t}$. This process can be formally represented as:

\begin{equation*}
act_t = \text{Questioner}_t(mem_t, env{t})
\end{equation*}
\begin{equation*}
qes_t = \text{Questioner}_t(act_t, plist{t})
\end{equation*}

Solver: The Solver processes the question generated by the Questioner and formulates a comprehensive plan, $plan_t$, comprising multiple actions. These actions, $act_t$, are designed to invoke appropriate tools and update the system's memory state. This process can be formally represented as:
\begin{equation*}
plan_t = \text{Solver}_t(qes_t, env_t)
\end{equation*}
\begin{equation*}
mem_{t+1}, act_t = \text{Solver}_t(plan_t)
\end{equation*}

\subsection{Dataset Generation Pipeline Overview}
\label{sec:pipeline_detail}

\begin{figure*}[htb!]
\centering
\includegraphics[width=\textwidth]{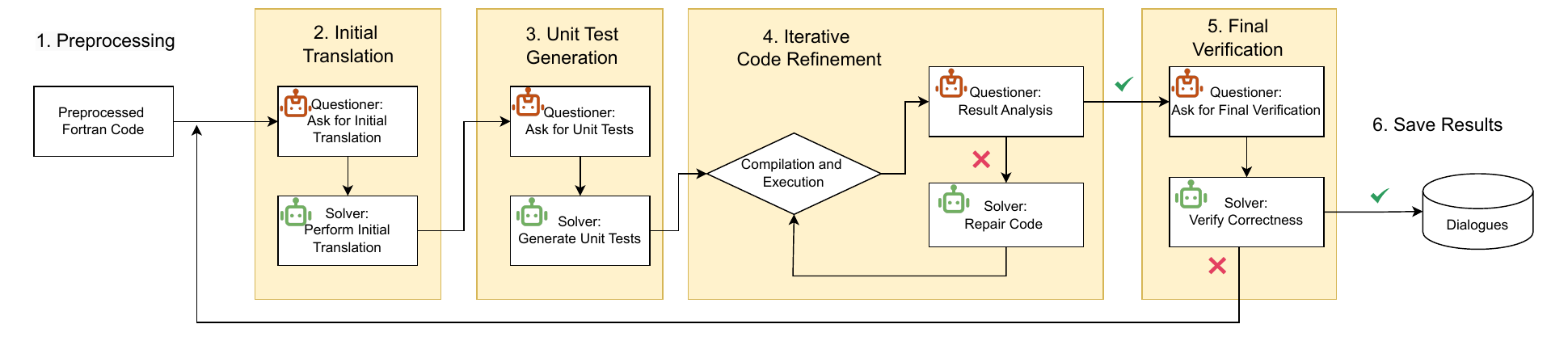} 
\caption{Overview of Dataset Generation Pipeline}
\label{fig:Overall_Pipeline}
\end{figure*}

Figure~\ref{fig:Overall_Pipeline} illustrates our dataset generation pipeline, including six phases: 
\textbf{preprocessing},
\textbf{initial translation}, \textbf{unit test generation}, \textbf{iterative code refinement}, \textbf{final verification}, and \textbf{saving results}. 
This approach enables specialized handling of various aspects of the translation process by integrating custom scripts and tools.

The specific details of each phase are discussed in Section~\ref{sec-generation-all}. Different from previous LLM agent implementations, our approach uses the proposed  Questioner-Solver module as the core instead of a single LLM. Section~\ref{sec:approach-module} introduces the  Questioner-Solver module in detail. The key components of our LLM agent system are as follows:

\noindent
\textbf{Agent core}: As shown in Figure~\ref{fig:agent}, the central module is responsible for managing the logic and behavioral characteristics of the agent. Unlike previous implementations that relied on a single LLM, our approach utilizes a questioner-solver module as the core. 

\noindent
\textbf{Memory}: Memory stores the agent’s internal logs and user interactions, tracking past decisions, actions, execution feedback, and observations. It facilitates the iterative refinement of translated code. Each iterative refinement approach carries a long-term memory to retain previous mistakes and solutions. The full memory history is saved and parsed to generate our muti-turn dialogue dataset.

\noindent
\textbf{Tools:} In our approach, tools refer to specialized external utilities that extend the agents' capabilities beyond language generation. Specifically, we use compilers (gfortran and g++) for code compilation, shell commands for execution, and custom Python scripts for tasks such as code parsing and analysis.

\noindent
\textbf{Environment:} This component refers to the data and information that the agent perceives or collects. It includes tool feedback, such as error messages and execution validation results, along with the current state of the translation.

\noindent
\textbf{Planning:} As shown in Figure~\ref{fig:Overall_Pipeline}, our pipeline refines the LLM-translated code through iterative loops. This process involves four major phases: step 2 through 5 represented by yellow blocks in Figure~\ref{fig:Overall_Pipeline}. Each phase devises a structured sequence of actions or steps to complete its assigned subtask. Table~\ref{tab:acts} lists the actions defined in our approach, along with their inputs, outputs, and invoked tools.

\begin{table*}[]
\caption{Actions Carried Out by the Solver.}
\resizebox{\textwidth}{!}{%
\begin{tabular}{llll}
\hline
Action                    & Input                                      & Output                                         & Envoriment   \\ \hline
Translate                 & Fortran Code                               & CPP code                                       & None         \\
Generate Test Cases       & Fortran or CPP Code                        & Fortran or CPP code with integrated test cases & None         \\
Compilation Fixing        & Fortran or CPP Code with errors            & A success or compilation  error message        & gcc compiler \\
Execution Fixing          & Fortran or CPP Code with errors            & A success or execution  error message          & CLI          \\
Inspect Test Case Results & Fortran or CPP Code with failed test cases & An updated code                                & CLI          \\
Keep Consistenty          & Fortran-CPP Code Pairs                     & Verified Fortran-CPP Code Pairs                & None         \\ \hline
\end{tabular}
}
\label{tab:acts}
\end{table*}

\subsection{LLM Agent-based Dataset Generation}
\label{sec-generation-all}

\subsubsection{Preprocessing} Selected input Fortran codes will go through a data preprocessing step to removes comments and checks for external dependencies. Removing comments from Fortran code before translation ensures that only executable logic is preserved. This helps eliminate unnecessary information, keep the input concise, reduce token consumption, and prevent outdated or irrelevant comments from interfering with code conversion.

Fortran code containing undefined external references will be skipped by this step. If the code relies on external dependencies that are not provided in the snippet, the translation to C++ may be incomplete or incorrect.

We also apply two filtering criteria: limiting token count to less than than a threshold (e.g. 600) to fit within LLM's input context window and including only executable code to facilitate unit testing. 

\subsubsection{Initial Translation} The next phase in our pipeline is to use LLMs to generate an initial translation given a Fortran code. 
The translation process from Fortran to C++ follows a structured interaction between a \textbf{Questioner} and a \textbf{Solver}. The Questioner first requests a translation, and the Solver provides the corresponding C++ code. Below are the prompts used in this process.

The Questioner provides the Fortran code and asks for a translation:

\begin{lstlisting}
q_ask_s_translation = """
Here is my Fortran code: {fortran_code}. 
Now you need to provide a complete question (including code) to the answerer and ask him to translate this Fortran code to C++. Don't translate this Fortran code by yourself.
"""
\end{lstlisting}

Upon receiving this request, the Solver generates the corresponding C++ translation:

\begin{lstlisting}
Prompts_Fortran_to_Cpp = """
Help me to translate the following Fortran code to C++ code: 
Fortran Code: {Fortran_Code}
Translated C++ Code:
"""
\end{lstlisting}

\subsubsection{Unit Test Generation} 
In this step, the Questioner-Solver module processes a pair of Fortran and C++ code to develop and integrate functionally equivalent unit test cases. This process is essential for ensuring behavioral consistency across both versions.

This step involves structured interaction between a \textbf{Questioner} and a \textbf{Solver}. The Questioner requests unit tests for both Fortran and C++ code, while the Solver generates the necessary test cases.

The Questioner asks for unit tests:

\begin{lstlisting}
q_ask_s_unit_test = """
Here is the answer from the solver: {ser_answer}, 
you now need to ask the answerer to provide the executable unit tests  for both the original Fortran code and the translated C++ code.  Please add the unit tests to the main function.  
In C++ code, you should use 'assert' for the unit test checking. One example: 
assert(has_close_elements((1.0, 2.0, 5.9, 4.0, 5.0), 0.95) == true).
In Fortran code, you should use the following format for the unit test checking:     
if (has_close_elements(a, n, 0.8)) then
    write(*,*) "Test case 2 failed: assertion failed"
    call exit(1)
end if
"""
\end{lstlisting}

These prompt are designed to ensure a structured and automated unit testing process. It enforces a structured test format, using `assert' in C++ and conditional checks in Fortran. The Solver is instructed not to use external testing frameworks, ensuring compatibility in different execution environments. These unit tests are generated in a way that allows them to be compiled and executed automatically as part of the translation workflow. Upon receiving this request, the Solver generates unit test code. 

By integrating unit tests into the translation process, this step ensures that the generated C++ code maintains functional equivalence with the original Fortran implementation.

\subsubsection{Iterative Refinement}

We use an iterative refinement step to verify correctness of the translation and repair any errors if possible. This step relies on feedback from compilation and execution of both Fortran and C++ programs. 
A python program is created as the harness to implement the logic of the workflow including compiling and running code, collecting errors, interacting with LLMs to fix errors.

If the Fortran or C++ code fails to compile, the error messages are extracted and passed into a Questioner prompt like the following to repair the code:

\begin{lstlisting}
Compiler_check_prompt = """
The compiler is throwing errors. The error report is: {reason}.  Please help me to continue modifying the C++ code.  Just write out the modified C++ code based on the error report.
New C++ Code:
"""
\end{lstlisting}

Similarly, if the execution of either program fails, the results are analyzed, and a correction is requested. Once a fix is suggested, the updated code is inserted into the iterative refinement pipeline. If errors persist after multiple iterations, the code is rejected. If errors disappear, the workflow enters the final verification step.

\subsubsection{Final Verification}
\label{sec-approach-consistency}

A final check is initiated by the Questioner to determine if the translation is functionally equivalent, using the following prompt:

\begin{lstlisting}
ft_ct_further_check = """
Fortran code outputs: {fortran_compile_run_result} 
C++ code result: {cpp_compile_run_result}
Based on the unit test results, tell me if the translation has been done correctly.
Just answer "Yes" or "No".
"""
\end{lstlisting}

If the response is "Yes", the final translated code is stored. If "No", the workflow goes back to Step 2.  In either case, the entire dialogue is stored into a dialogue history. 

\subsection{Fortran2Cpp Datasets}
In the end, the outlined translation approach can be used to generate two types of datasets: a paired Fortran-C++ dataset and a dialogue dataset. By adjusting the size of the seed Fortran codes, the entire pipeline can be scaled to produce as much data as required, with the only constraints being resources and time.

For example, we sourced Fortran code from CodeParrot's GitHub repository \cite{codeparrot_dataset}, which contains 115 million code files across 32 programming languages. Due to resource limitations, we selected the first 80,000 of the 142,038 available Fortran files as our seed input.


\paragraph{Code Pair Dataset}: using the proposed pipeline, we generated 2,529 Fortran-C++ data pairs with a successful data conversation rate at 29.6\%. Each of the paired data has gone through the compilation and execution verification in our pipeline. 

\paragraph{Multi-Dialogue Dataset}
We first give two definitions following the work of \citet{yi2024survey}.
\begin{itemize}[noitemsep, topsep=0pt, leftmargin=*]
    \item Dialogue: A complete sequence of interactive communication between two or more agents, with a clear beginning and end, unified by a common context or purpose. It's composed of multiple dialogues/turns and represents the full scope of the interaction.
    \item Turn: A single turn of exchange between agents, consisting of one utterance/message and its corresponding response. This forms the basic unit of interaction within a dialogue.
\end{itemize}


We split a multi-turn dialogue into multiple dialogues based on the number of turns the original dialogue has. 
For example, a three-turn dialogue (labeled \( s_0 \) through \( s_3 \)) is structured into three cumulative prompt-response pairs to maintain contextual continuity. Starting with the initial prompt \( p_1 \), containing only the first turn (\( s_0 \)), we progressively expand each subsequent prompt by incorporating all preceding turns. Specifically, \( p_2 \) consists of the accumulated dialogue up to \( s_1 \) (i.e., \( s_0 + s_1 \)) as the prompt, generating response \( s_2 \); and \( p_3 \) includes \( s_0 + s_1 + s_2 \) as the prompt, generating \( s_3 \). By this final step, the model processes the entire preceding dialogue, ensuring continuity and refinement in the generated responses.

Translation dialogue datasets capture semantics from both static source code and its dynamic interactions with compilers and runtime environments, embedding knowledge of compiler diagnostics, error resolutions, and iterative refinements. Training LLMs on these workflows enhances their understanding of code evolution, improving translation accuracy and enabling preemptive error correction for more robust outputs. This approach also maximizes data efficiency by leveraging intermediate steps instead of discarding erroneous attempts, ensuring models learn from the full translation process while reducing reliance on excessive raw code pairs.

Due to the resource limit, we selected 1.2K dialogues between the Questioner and Solver and split them to multiple prompt-response pairs. 
After splitting after each turn, a history of 1.2k dialogues generates 11.7k prompt-response pairs.





\section{Experiment} 


\label{sec-experiments}
In this section, we present the experimental setup and results of using Fortran2CPP fine-tuning selected LLMs for Fortran to C++ translation. 

\subsection{Experiment Setup}
\begin{table*}[ht]
\small
\centering
\caption{Selected Code-Oriented Language Models} 
\label{tab:selected-models}
\begin{adjustbox}{max width=0.85\textwidth}
\begin{tabular}{l|c|c|c|c}
\toprule
\textbf{Specification} & \textbf{DeepSeek-Coder} & \textbf{CodeLlama} & \textbf{StarCoder} & \textbf{GPT-4 Turbo} \\
\midrule
Parameters & 6.7B & 13B & 15.5B & Unknown \\
Training Data & 2T tokens & 500B tokens & 250B tokens & Unknown \\
Context Window & 16K & 100K & 8K & 128K \\
Open Weights & Yes & Yes & Yes & No \\
Developer & DeepSeek AI & Meta & BigCode & OpenAI \\
\bottomrule
\end{tabular}
\end{adjustbox}
\end{table*}

\noindent
\textbf{Models:}
As shown in Table~\ref{tab:selected-models}, we select a few representative LLMs to evaluate our Fortran2CPP dataset, including DeepSeek-Coder~\cite{guo2024deepseek}, Codellama~\cite{roziere2023code}, StarCoder~\cite{li2023starcoder}, and GPT-4-Turbo~\cite{gpt-4-turbo}.  These models represent a range of sizes and training methodologies, providing a comprehensive assessment of LLM capabilities for code translation. 

\noindent
\textbf{Evaluation Datasets:} The evaluation employs two distinct datasets to assess LLMs' Fortran to C++ translation capabilities. The first is the HPC-Fortran-Cpp dataset~\cite{lei2023creating}, comprising 315 manually curated pairs of OpenMP Fortran and C++ code. Due to a 4000-token limit on Fortran source code, 296 pairs were ultimately selected for the translation task.
The second dataset builds upon HumanEval-X~\cite{zheng2023codegeex}, which originally evaluated code generation models across five programming languages (Python, C++, Java, JavaScript, and Go) using 164 test cases. We expanded this dataset by creating corresponding Fortran implementations for the C++ examples. Using a GPT-4-based pipeline, followed by iterative compilation, execution, and refinement, we generated 126 validated pairs of matching Fortran and C++ code snippets.

\noindent
\textbf{Evaluation Metrics:}
The quality of the translated C++ code is assessed using four complementary metrics:
\begin{itemize}
    \item \textit{CodeBLEU Score:} measures C++ translation similarity (0-1.0) using n-gram, syntax, and dataflow components. 
    \item \textit{Compilation Check:} determines whether the generated C++ code is syntactically valid by attempting compilation with the GNU C++ compiler, yielding a success rate between 0 and 1.0.
    \item \textit{Execution Test:} verifies functional correctness by running the translated code against unit tests generated by GPT-4, measuring the proportion of matching outputs between Fortran and C++ implementations.
    \item \textit{Manual Investigation:}  involves expert review of randomly selected samples (30 from HPC-Fortran-C++ and 20 from HumanEval-Fortran) scored on a 0-5 scale based on translation accuracy and completeness. The detailed grading criteria are provided in Table~\ref{tab:grading}.
\end{itemize}

\noindent
\textbf{Implementation Details:}
Experiments were conducted on an Nvidia A100 (80GB) GPU using Hugging Face Transformers~\cite{huggingfaceTransformers} for model inference, with temperature set to 0.2. This setup enables comprehensive evaluations of LLM performance in Fortran to C++ translation across our dataset and metrics.

\noindent
\textbf{Hyper-parameters:} The fine-tuning configuration employed key hyperparameters, including a conservative learning rate of 9.65e-6, weight decay of 0.1, and sequence length limit of 1024 tokens. We fine-tuned the model for 3 epochs using a cosine learning rate scheduler without warmup steps. 


\subsection{Results and Analysis}

\textbf{RQ1: Effectiveness of the Fortran2CPP Dataset for Fine-tuning LLMs.} 
We fine-tuned three open-weight LLMs, DeepSeek-Coder (6.7B), CodeLlama (13B), and StarCoder (15.5B), on the Fortran2CPP dataset and evaluated their performance on the HPC-Fortran-Cpp and HumanEval-Fortran2Cpp datasets. 

Tables~\ref{tab:HPC-FORTRAN-CPP} shows the evaluation results on the HPC-Fortran-Cpp dataset. We observe significant improvements across all evaluation metrics. Notably, Codellama 13B gains a substantial 1.79x increase in CodeBLEU score (from 0.09 to 0.16) as shown in Table~\ref{tab:codeblue-ratio-datasets}. The execution test ratio for DeepSeek-Coder 6.7B increased from 0 to 0.65. 

Table~\ref{tab:Humaneval-F2C++} showcases the performance of the same fine-tuned LLMs evaluated on the HumanEval-Fortran2Cpp dataset. Similarly, we observe noticeable improvements in most metrics after fine-tuning. For example, StarCoder 15.5B achieves a substantial 3.31x increase in CodeBLEU score (from 0.09 to 0.24) as shown in Table~\ref{tab:codeblue-ratio-datasets}. The execution test ratio for Codellama 13B increased from 0 to 0.92. 

These results highlight the importance of specialized datasets for improving the performance of LLMs in specific code translation tasks. It also suggests that fine-tuning on a larger and more diverse dataset like the Fortran2Cpp Dataset can be generalized to improve performance on smaller, more challenging benchmarks like HumanEval-Fortran2Cpp. 
 


\textbf{RQ2: Value of Multi-turn Dialogue Integration.} Our ablation study compared models fine-tuned solely on Fortran-C++ code pairs against those trained with Fortran2CPP dialogue data capturing error correction and iterative refinement. Models trained with the dialogue-enhanced dataset consistently outperformed their counterparts across all metrics, as shown in Table~\ref{tab:HPC-FORTRAN-CPP} and~\ref{tab:Humaneval-F2C++}. Table~\ref{tab:codeblue-ratio-datasets} also shows the same trend in the CodeBLEU ratio values. These findings underscore the importance of contextual learning and demonstrates how dialogue data helps models understand both the translation process and error correction strategies.

\textbf{RQ3: Comparison with State-of-the-art Models}. When compared to GPT-4 Turbo, our fine-tuned CodeLlama demonstrated superior performance in CodeBLEU score (0.239) and compilation success rate (0.93). However, GPT-4 Turbo maintained an edge in execution test success rate (0.83 versus 0.74), suggesting that while our approach achieves competitive results in structural accuracy and syntactic correctness, there remains room for improvement in ensuring functional equivalence. These results, detailed in  Table~\ref{tab:HPC-FORTRAN-CPP} and~\ref{tab:Humaneval-F2C++}, indicate that our fine-tuned models can match or exceed state-of-the-art performance on specific metrics while maintaining competitive performance overall.

\begin{table*}[htbp]
  \centering
  \caption{Performance of different fine-tuned models on the HPC-Fortran-Cpp dataset.}
  \label{tab:HPC-FORTRAN-CPP}
  \begin{adjustbox}{width=0.9\textwidth}
  \begin{tabular}{@{}l|c|ccccc@{}}
    \toprule
    \textbf{Evaluation Method} & \textbf{Before/After Fine-tuning} & \textbf{DeepSeek-Coder 6.7B} & \textbf{CodeLlama 13B} & \textbf{StarCoder 15.5B} & \textbf{GPT-4 Turbo} \\ 
    \midrule
    \multirow{3}{*}{CodeBLEU Score} 
      & Original & 0.096 & 0.090 & 0.092 & \textbf{0.262} \\ 
      & Fine-tuned with Code Pairs & 0.108 & 0.101 & 0.103 & \\ 
      & Fine-tuned with Dialogue & {0.149} & {0.161} & {0.159} & \\ 
    \addlinespace
    \midrule
    \multirow{3}{*}{Compilation Check} 
      & Original & 0.0 (0 / 296) & 0.0 (0 / 296) & 0.0 (0 / 296) & \textbf{0.70 (207 / 296)} \\ 
      & Fine-tuned with Code Pairs & 0.62 (183 / 296) & 0.44 (131 / 296) & 0.36 (107 / 296) & \\ 
      & Fine-tuned with Dialogue & \textbf{0.70 (207 / 296)} & {0.67 (199 / 296)} & {0.69 (204 / 296)} & \\ 
    \addlinespace
    \midrule
    \multirow{3}{*}{Execution Test Evaluation} 
      & Original & 0.0 (0 / 296) & 0.0 (0 / 296) & 0.0 (0 / 296) & 0.48 (143 / 296) \\ 
      & Fine-tuned with Code Pairs & 0.55 (162 / 296) & 0.37 (109 / 296) & 0.33 (98 / 296) & \\ 
      & Fine-tuned with Dialogue & \textbf{0.65 (191 / 296)} & {0.54 (160 / 296)} & {0.52 (155 / 296)} & \\ 
    \addlinespace
    \midrule
    \multirow{3}{*}{Manual Investigation} 
      & Original & 0 & 0 & 0.37 & 4.4 \\ 
      & Fine-tuned with Code Pairs & 3.7 & 2.7 & 3.37 & \\ 
      & Fine-tuned with Dialogue & \textbf{4.5} & 4.03 & \textbf{4.5} & \\ 
    \bottomrule
  \end{tabular}
  \end{adjustbox}
\end{table*}

\begin{table*}[ht]
  \centering
  \caption{Performance of different fine-tuned models on HumanEval-Fortran2Cpp.}
  \label{tab:Humaneval-F2C++}
  \begin{adjustbox}{width=0.9\textwidth}
  \begin{tabular}{@{}l|c|ccccc@{}}
    \toprule
    \textbf{Evaluation Method} & \textbf{Before/After Fine-tuning} & \textbf{DeepSeek-Coder 6.7B} & \textbf{CodeLlama 13B} & \textbf{StarCoder 15.5B} & \textbf{GPT-4 Turbo} \\ 
    \midrule
    \multirow{3}{*}{CodeBLEU Score} 
      & Original & 0.072 & 0.090 & 0.067 & 0.203 \\ 
      & Fine-tuned with Code Pairs & 0.098 & 0.101 & 0.089 & \\ 
      & Fine-tuned with Dialogue & 0.225 & \textbf{0.239} & 0.225 & \\ 
    \addlinespace
    \midrule
    \multirow{3}{*}{Compilation Check} 
      & Original & 0.0 (0 / 126) & 0.0 (0 / 126) & 0.0 (0 / 126) & 0.90 (113 / 126) \\ 
      & Fine-tuned with Code Pairs & 0.64 (81 / 126) & 0.60 (75 / 126) & 0.59 (74 / 126) & \\ 
      & Fine-tuned with Dialogue & 0.84 (106 / 126) & \textbf{0.92 (116 / 126)} & 0.64 (81 / 126) & \\ 
    \addlinespace
    \midrule
    \multirow{3}{*}{Execution Test Evaluation} 
      & Original & 0.0 (0 / 126) & 0.0 (0 / 126) & 0.0 (0 / 126) & \textbf{0.83 (104 / 126)} \\ 
      & Fine-tuned with Code Pairs & 0.61 (77 / 126) & 0.46 (58 / 126) & 0.43 (54 / 126) & \\ 
      & Fine-tuned with Dialogue & 0.71 (89 / 126) & 0.74 (93 / 126) & 0.51 (64 / 126) & \\ 
    \addlinespace
    \midrule
    \multirow{3}{*}{Manual Investigation} 
      & Original & 0 & 3.75 & 0 & \textbf{4.75} \\ 
      & Fine-tuned with Code Pairs & 3.75 & 3.2 & 3.4 & \\ 
      & Fine-tuned with Dialogue & 4.7 & \textbf{4.75} & 4.7 & \\ 
    \bottomrule
  \end{tabular}
  \end{adjustbox}
\end{table*}

\begin{table}[ht]
\caption{CodeBLEU Ratio (fine-tuned/original) for selected datasets: HPC-Fortran-Cpp(HPC), HumanEval-Fortran2Cpp(HumanEval)}
\label{tab:codeblue-ratio-datasets}
\centering
\begin{adjustbox}{max width=\linewidth}
\begin{tabular}{>{\bfseries}l >{\bfseries}l cc}
\toprule
Model & Fine-tuned & \multicolumn{2}{c}{\textbf{CodeBLEU Ratio}} \\
      & with      & \multicolumn{2}{c}{} \\
\cmidrule(l){3-4}
      &           & \textbf{HPC-Fortran-Cpp} & \textbf{HumanEval} \\
\midrule
\multirow{2}{*}{DeepSeek-Coder} & Code Pairs & 1.13 & 1.36 \\
                                & Dialogue   & 1.55 & 3.11 \\
\cmidrule(lr){1-4}
\multirow{2}{*}{Codellama}      & Code Pairs & 1.11 & 1.12 \\
                                & Dialogue   & 1.79 & 2.64 \\
\cmidrule(lr){1-4}
\multirow{2}{*}{StarCoder}      & Code Pairs & 1.12 & 1.33 \\
                                & Dialogue   & 1.73 & 3.31 \\
\bottomrule
\end{tabular}
\end{adjustbox}
\end{table}

\section{Related Work}

Fine-tuning large language models (LLMs) for programming language translation has shifted from general-purpose models like GPT-2 and GPT-3, which often produced incomplete translations, to approaches specifically tailored for this task \cite{chen2021evaluating}. Models such as Codex and PolyCoder, built on the GPT architecture and enhanced with large programming datasets, improve translation accuracy but still face challenges with differing language paradigms \cite{xu2022systematic,chen2021evaluating}.
Transfer learning enhances translation by pre-training on a source language and fine-tuning on a target, leveraging structural similarities yet facing limitations with under-resourced languages and noisy data \cite{ahmad2021unified}. TransCoder employs a self-supervised, bidirectional approach for unsupervised translation, effective for popular languages but less so for obscure ones due to data scarcity \cite{roziere2020unsupervised}.

The scarcity of parallel corpora remains a significant challenge, with recent efforts focused on creating large-scale multilingual datasets. CodeSearchNet, while widely used, lacks parallelism for direct translation tasks \cite{husain2019codesearchnet}. Techniques like automatic mining have been explored to address this, as seen in TRANX \cite{yin2018tranx} and the BigCode Project \cite{allamanis2018survey}.
Datasets like MCoNaLa \cite{wang2022mconala}, which pairs natural language with code, and CodeParrot \cite{codeparrot_dataset}, focused on high-quality code snippets, contribute to enhancing LLM training. Additionally, self-supervised learning and back-translation techniques generate parallel data, using round-trip consistency for model improvement \cite{roziere2020unsupervised}.

\section{Conclusion}
This paper addresses the significant challenges inherent in Fortran-to-C++ translation by introducing an innovative LLM-based approach featuring a novel Questioner-Solver module. This approach not only generates a valuable multi-turn dialogue dataset to enhance LLMs' domain knowledge but also demonstrates substantial improvements across multiple metrics, affirming its effectiveness in increasing syntactic accuracy and functional reliability. The results highlight the potential of dialogue-based LLM training to revolutionize legacy code modernization, proposing a new paradigm for automated code translation technologies. Future efforts will focus on expanding dataset diversity and enhancing model capabilities to advance HPC translation tasks.

\section{Limitations}
The translation approach in this work relies on LLMs to generate unit tests to ensure the logical correctness and consistency between the source and translated code. While previous research~\cite{chen2024chatunitest, schafer2023empirical} has successfully leveraged LLMs to generate unit test cases, adopting a stricter validation process for this step would be beneficial.

\clearpage
\bibliography{main} 

\appendix

\section{Appendix}
\label{sec:appendix}

\subsection{Fortran2Cpp Dataset Analysis}
\begin{figure*}[hbt]
    \centering
    \begin{subfigure}[b]{0.45\textwidth}
        \centering
        \includegraphics[width=\columnwidth]{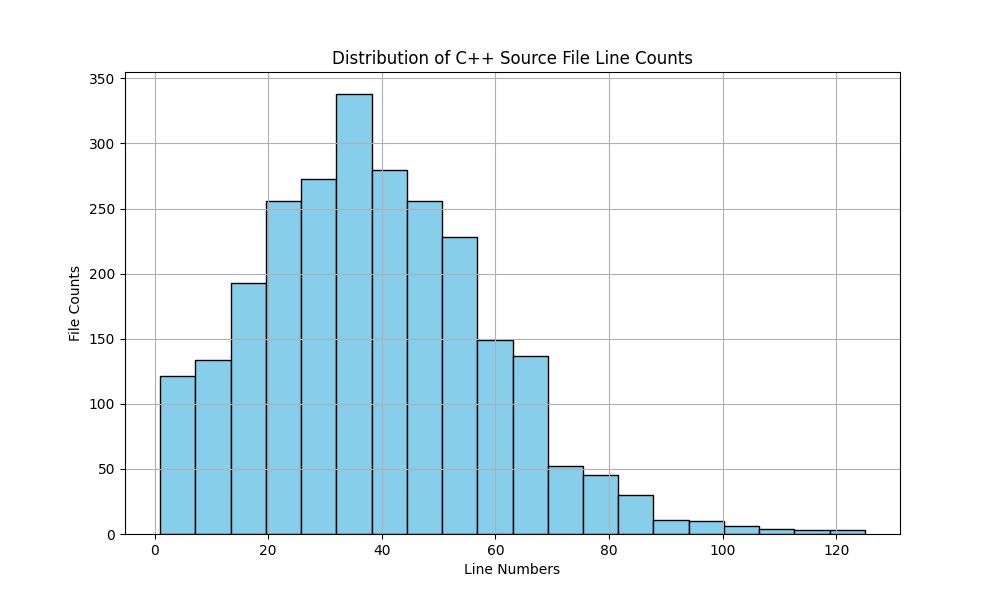} 
\caption{Distribution of C++ Source File Line Counts.}
\label{fig:C++_distribution}
    \end{subfigure}
    \hfill
    \begin{subfigure}[b]{0.45\textwidth}
        \centering
        \includegraphics[width=\columnwidth]{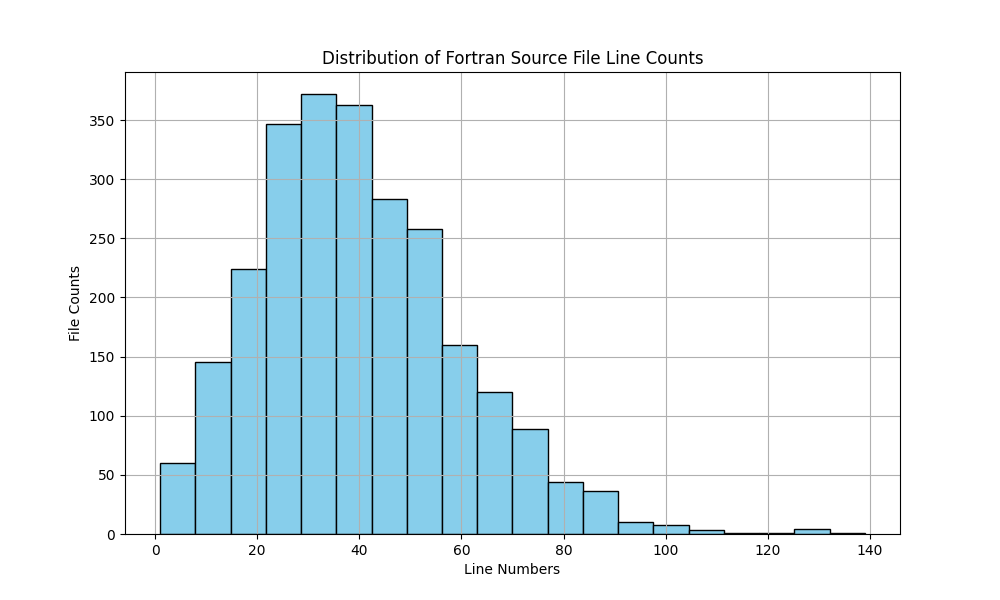} 
        \caption{Distribution of Fortran Source File Line Counts.}
        \label{fig:fortran_distribution}
    \end{subfigure}
    \caption{Comparison of C++ and Fortran source file line count distribution.}
    \label{fig:keywords_comparison1}
\end{figure*}

\begin{figure*}[hbt]
    \centering
    \begin{subfigure}[b]{0.48\textwidth}
        \centering
        \includegraphics[width=\textwidth]{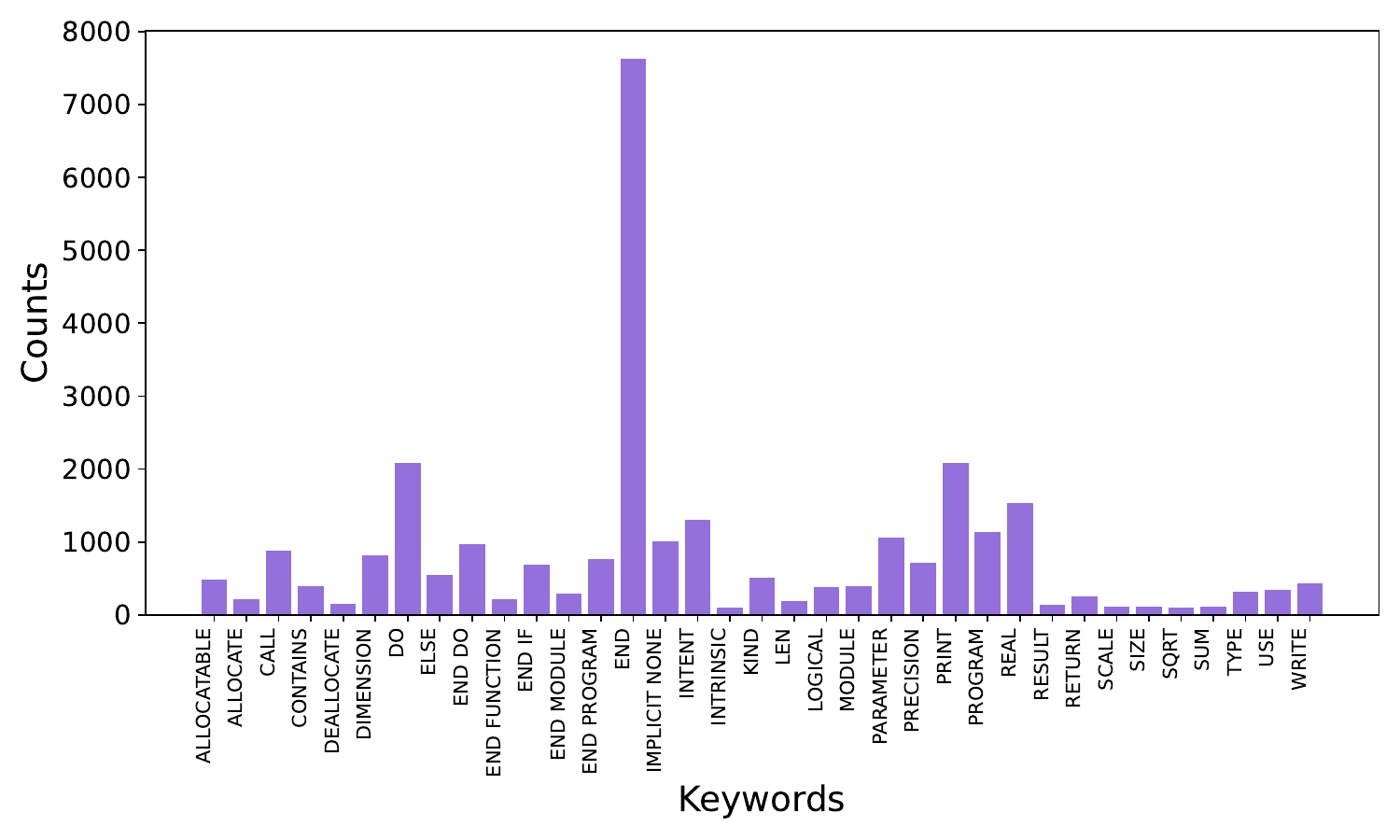}
        \caption{Fortran keywords histogram.}
        \label{fig:fortran_keyword}
    \end{subfigure}
    \hfill
    \begin{subfigure}[b]{0.48\textwidth}
        \centering
        \includegraphics[width=\textwidth]{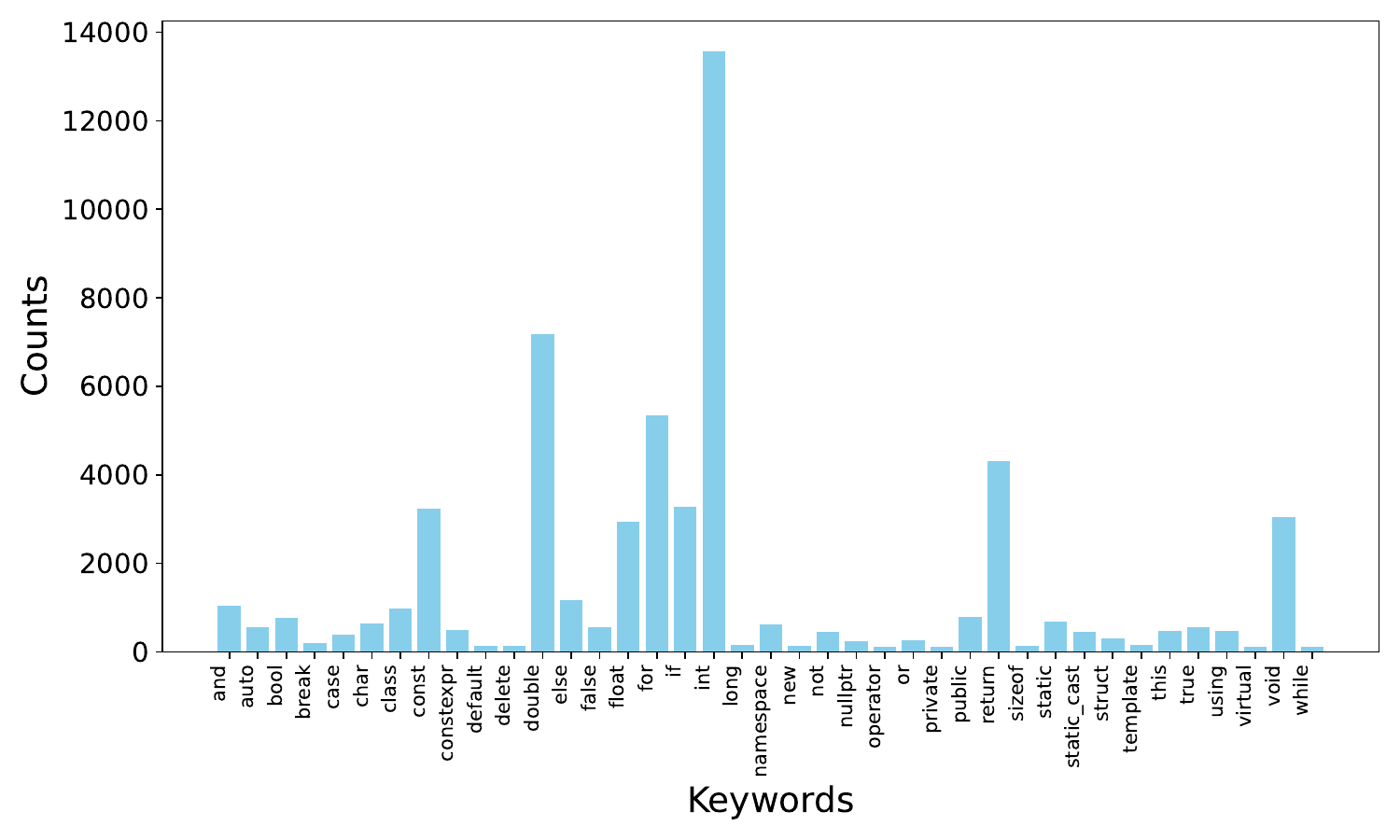}
        \caption{C++ keywords histogram}
        \label{fig:cpp_keywords}
    \end{subfigure}
    \caption{Comparison of C++ and Fortran keyword histograms.}
    \label{fig:keywords_comparison2}
\end{figure*}

Figure~\ref{fig:keywords_comparison1} and Figure~\ref{fig:keywords_comparison2} show the distribution of Fortran and C++ keywords in Fortran2CPP's code pair dataset.

\begin{figure*}[H]
    \centering
    \begin{subfigure}[b]{0.48\textwidth}
        \centering
        \includegraphics[width=\textwidth]{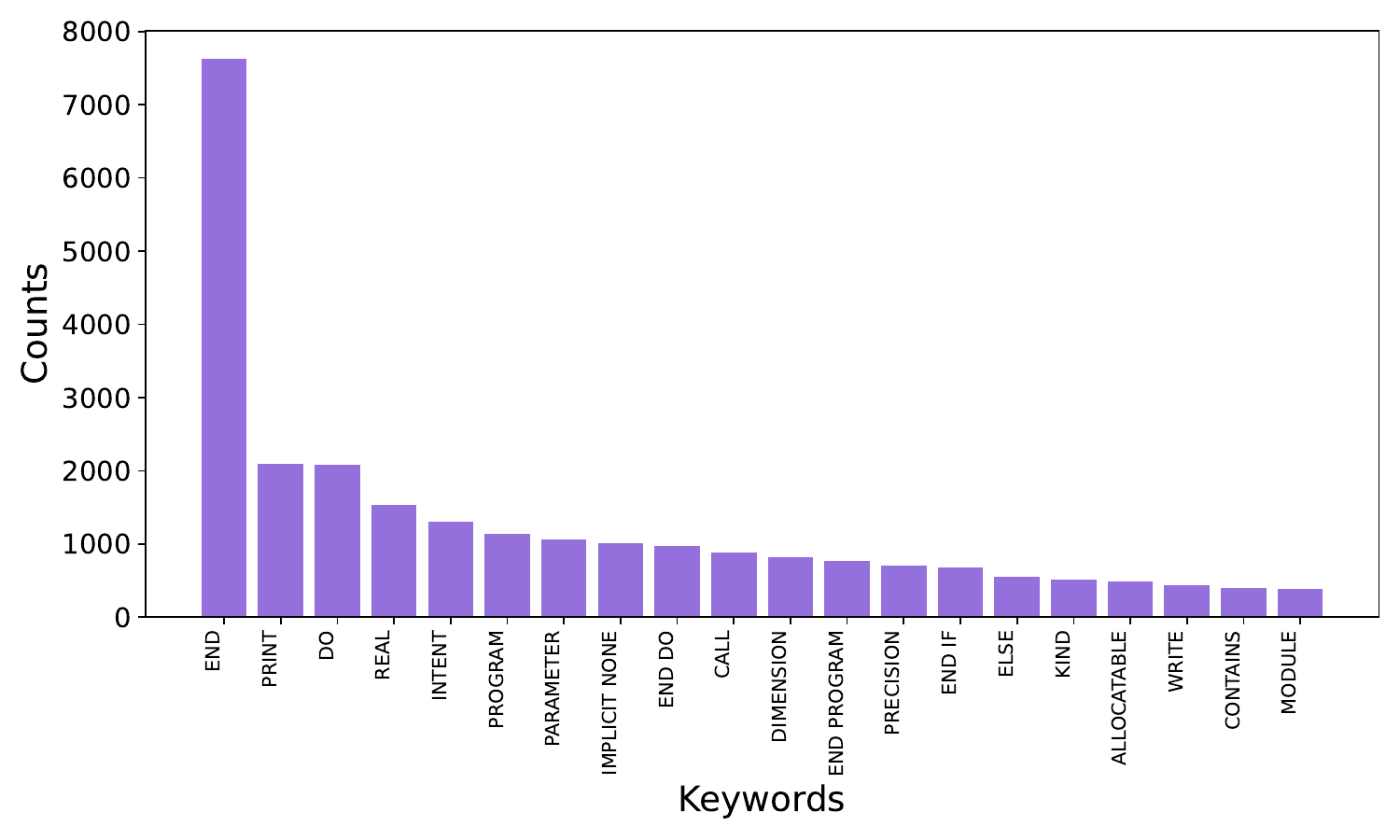} 
\caption{Fortran top20 keywords histogram.}
        \label{fig:fortran_top20}
    \end{subfigure}
    \hfill
    \begin{subfigure}[b]{0.48\textwidth}
        \centering
        \includegraphics[width=\textwidth]{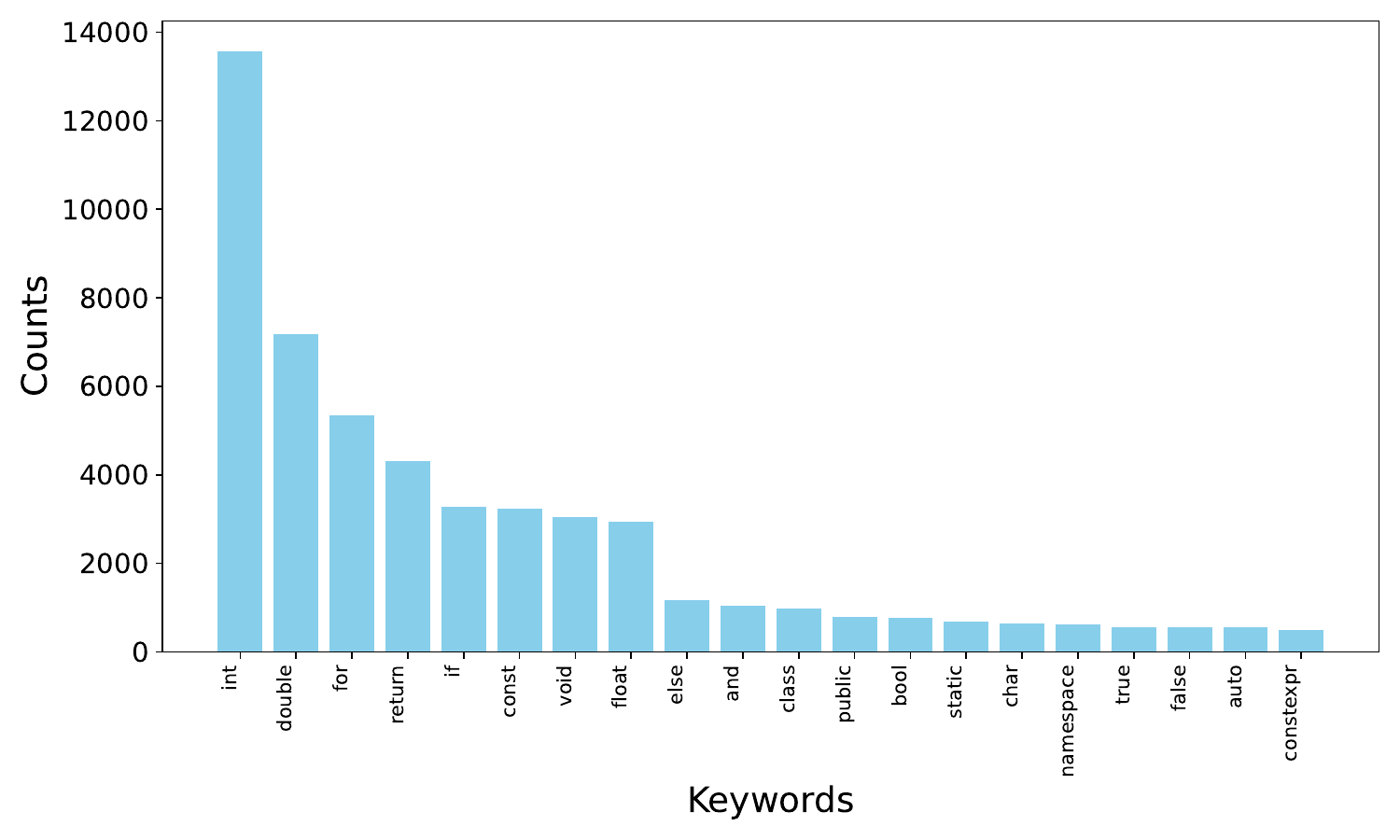} 
\caption{C++ top20 keywords histogram}
        \label{fig:cpp_top20}
    \end{subfigure}
    \caption{Comparison of C++ and Fortran keyword histograms.}
    \label{fig:keywords_comparison3}
\end{figure*}

\subsection{Manual Grading Metrics}
Random sample assessment (30 HPC-Fortran-C++, 20 HumanEval-Fortran entries) scored 0-5 on translation accuracy and completeness. Details of manual grading metrics are listing in Table~\ref{tab:grading}.

\begin{table*}[h]
    \caption{Manual Grading of ML-Translated Fortran to C++ Code}
    \label{tab:grading}
    \centering
    \begin{tabular}{|c|p{5cm}|p{9.5cm}|}
        \hline
        \textbf{Score} & \textbf{Definition} & \textbf{Example} \\
        \hline
        0 & Irrelevant Translation with no recognizable C++ output & 
        \scriptsize\begin{verbatim}
//incorrect: output is same as the input Fortran code
function sum_array(arr, n) result(sum)
    integer, intent(in) :: n
    integer, dimension(n), intent(in) :: arr
    integer :: sum, i

    sum = 0
    do i = 1, n
        sum = sum + arr(i)
    end do
end function sum_array            
        \end{verbatim}
         \\
        \hline
        1 & Highly Incomplete Translation with only basic C++ code structure, such as function structure (name and body), translated & 
        \scriptsize\begin{verbatim}
//incorrect: only function body structure is present

    int sum_array(int arr[], int n) {
    return result;
}        
        \end{verbatim}
         \\
         \hline
        2 & Incomplete Translation with more translated but incorrect statements & 
    \scriptsize\begin{verbatim}
// incorrect: incorrect array syntax in argument and reference
            int sum_array(int arr, int n) {
    do (int i = 0; i < n; i++) {
        sum = sum + arr(0); 
    }
    return sum;
}
        \end{verbatim}
         \\
        \hline
        3 & Partial but Recognizable Translation but with major syntactic errors  & 
        \scriptsize\begin{verbatim}
// incorrect: using "do" instead of "for",
// and always accesses first element.
            int sum_array(int arr[], int n) {
    int sum = 0;
    do (int i = 0; i < n; i++) {
        sum = sum + arr[0]; 
    }
    return sum;
}
        \end{verbatim}
         \\
        \hline
        4 & Mostly Complete Translation with few incorrectly translated details, such as loop indices, or minor syntactic errors & 
        \scriptsize\begin{verbatim}
// incorrect: array bound incorrectly starts at 1, instead of 0.
int sum_array(int arr[], int n) {
    int sum = 0;
    for (int i = 1; i <= n; i++) {
        sum += arr[i];
    }
    return sum;
}
        \end{verbatim} \
         \\
        \hline
        5 & Complete Translation with details handled & 
         \scriptsize\begin{verbatim}
int sum_array(int arr[], int n) {
    int sum = 0;
    for (int i = 0; i < n; i++) {
        sum += arr[i];
    }
    return sum;
}
        \end{verbatim} 
         \\
        \hline
    \end{tabular}
\end{table*}

\subsection{Example of Splitting a Dialogue}

As a more concrete example, the following input JSON has a two-turn dialogue.  
\begin{lstlisting}
[
{
    "id": "conv1",
    "messages": [
        {"role": "user", "content": "Hi"},
        {"role": "assistant", "content": "Hello!"},
        {"role": "user", "content": "How are you?"},
        {"role": "assistant", "content": "I'm good, thank you."}
    ]   
}
]
\end{lstlisting}

The corresponding output after splitting is shown below.
\begin{lstlisting}
[
{
    "id": "conv1",
    "messages": [
        {"role": "user", "content": "Hi"},
        {"role": "assistant", "content": "Hello!"}
    ]   
},
{
    "id": "conv1",
    "messages": [
        {"role": "user", "content": "Hi"},
        {"role": "assistant", "content": "Hello!"},
        {"role": "user", "content": "How are you?"},
        {"role": "assistant", "content": "I'm good, thank you."}
    ]   
}
]

\end{lstlisting}

\end{document}